\documentclass{sig-alternate-05-2015}

\usepackage{times}
\usepackage{epsfig}
\usepackage{graphicx}
\usepackage{amsmath}
\usepackage{amssymb}
\usepackage{subcaption}

\usepackage[pagebackref=true,breaklinks=true,letterpaper=true,colorlinks,bookmarks=false]{hyperref}

\begin{document}

\title{YouTube-8M: A Large-Scale Video Classification Benchmark}

\author{Sami Abu-El-Haija\\
{\tt\small haija@google.com} \\
\and
Nisarg Kothari \\
{\tt\small ndk@google.com} \\
\and
Joonseok Lee \\
{\tt\small joonseok@google.com} \\
\and
Paul Natsev \\
{\tt\small natsev@google.com} \\
\and
George Toderici \\
{\tt\small gtoderici@google.com} \\
\and
Balakrishnan Varadarajan \\
{\tt\small balakrishnanv@google.com} \\
\and
Sudheendra Vijayanarasimhan \\
{\tt\small svnaras@google.com} \\
\and
Google Research \\
}

\maketitle

\begin{abstract}
Many recent advancements in Computer Vision are attributed to large
datasets. Open-source software packages for Machine Learning and inexpensive
commodity hardware have reduced the barrier of entry for exploring novel
approaches at scale.  It is possible to train models over millions of examples within a few days.
Although large-scale datasets exist for image understanding, such as ImageNet,
there are no comparable size video classification datasets.

In this paper, we introduce {\bf YouTube-8M}, the largest \emph{multi-label video
classification dataset}, composed of $\mathtt{\sim}8$ million videos---$500K$
hours of video---annotated with a vocabulary of $4800$ visual entities. To get
the videos and their (multiple) labels, we used a YouTube video annotation
system, which labels videos with the main topics in them. While the labels are
machine-generated, they have high-precision and are derived from a variety of
human-based signals including metadata and query click signals, so they
represent an excellent target for {\it content-based annotation} approaches. We
filtered the video labels (Knowledge Graph entities) using both automated and
manual curation strategies, including asking human raters if the labels are
{\it visually recognizable}. Then, we decoded each video at one-frame-per-second,
and used a Deep CNN pre-trained on ImageNet to extract the hidden representation
immediately prior to the classification layer. Finally, we compressed the frame
features and make both the features and video-level labels available for
download. The dataset contains frame-level features for over $1.9$ billion video
frames and $8$ million videos, making it the largest public multi-label video dataset.

We trained various (modest) classification models on the dataset, evaluated
them using popular evaluation metrics, and report them as
baselines. Despite the size of the dataset, some of our models train to
convergence in less than a day on a single machine using the publicly-available
TensorFlow framework. We plan to release code for training a basic
TensorFlow model and for computing metrics.

We show that pre-training on large data generalizes to other datasets like
Sports-1M and ActivityNet. We achieve state-of-the-art on ActivityNet, improving
mAP from $53.8\%$ to $77.6\%$. We hope that the unprecedented scale and diversity of YouTube-8M
will lead to advances in video understanding and representation learning.
\end{abstract}

\section{Introduction}
\label{sec:intro}

\begin{figure}[t]
\begin{center}
  \includegraphics[width=1.0\linewidth]{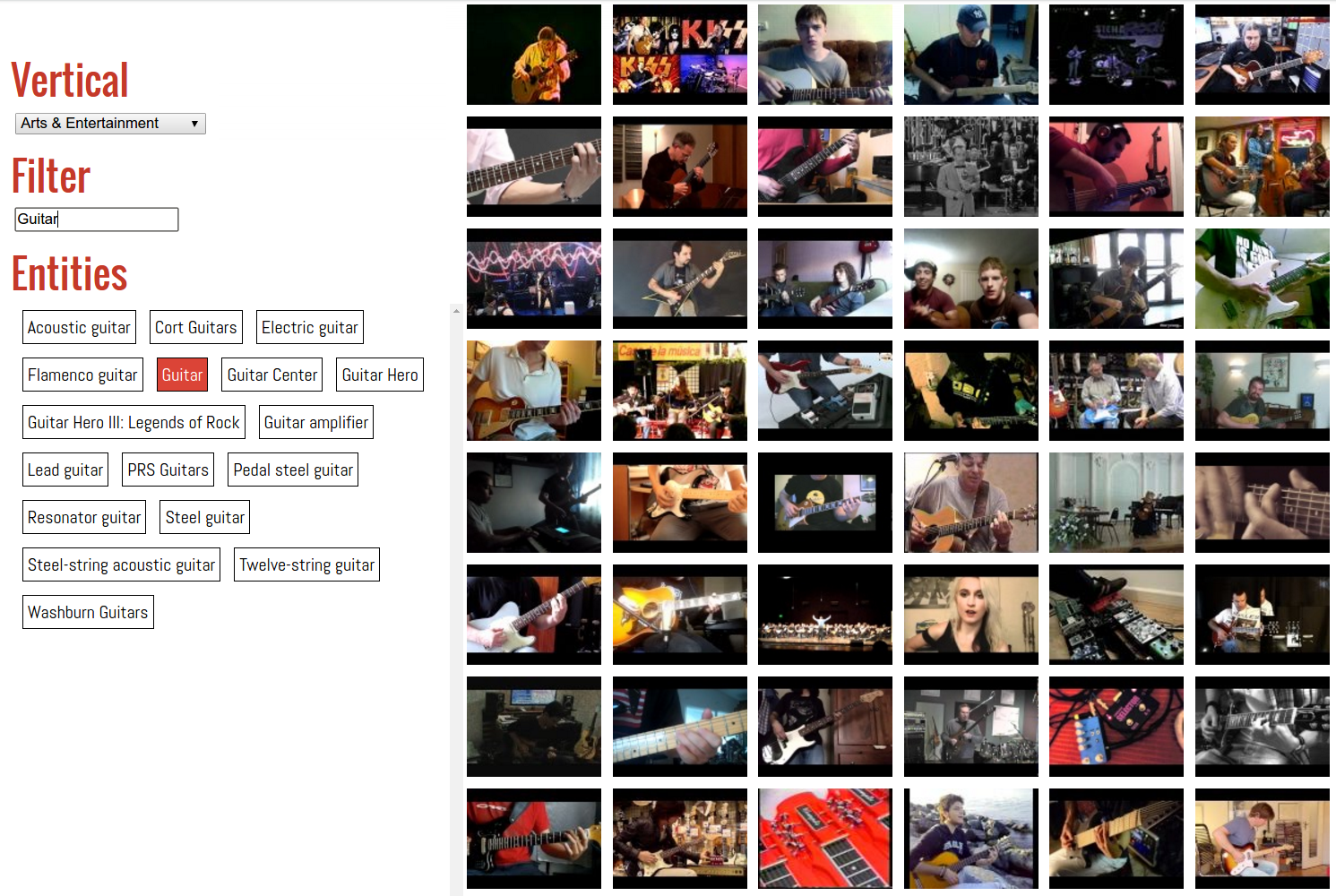}
\end{center}
\caption{YouTube-8M is a large-scale benchmark for general multi-label video classification. This screenshot of a dataset explorer depicts a subset of videos in the dataset annotated with the entity ``Guitar''.  The dataset explorer allows browsing and searching of the full vocabulary of Knowledge Graph entities, grouped in 24 top-level verticals, along with corresponding videos. 
         }
\label{fig:examples}
\end{figure}

Large-scale datasets such as ImageNet~\cite{ImageNet} have been key enablers of
recent progress in image understanding ~\cite{krizhevsky2012imagenet, inception-bn, resnet}.
By supporting the learning process of deep networks with millions of parameters, such datasets have played a crucial role
for the rapid progress of image understanding to near-human level accuracy~\cite{imagenetreview}.
Furthermore, intermediate layer activations of such networks have proven to be
powerful and interpretable for various tasks beyond
classification~\cite{oldzeiler13visualizing, girshick15fastrcnn, overfeat}.
In a similar vein, the amount and size of video benchmarks is growing with the
availability of Sports-1M~\cite{karpathy2014large} for sports videos and
ActivityNet~\cite{actnet} for human activities. However, unlike ImageNet, which
contains a diverse and general set of objects/entities, existing video benchmarks
are restricted to action and sports classes.

In this paper, we introduce YouTube-8M~\footnote{http://research.google.com/youtube8m}, a large-scale benchmark dataset for general
\emph{multi-label video classification}. We treat the task of video
classification as that of producing labels that are relevant to a video given
its frames. Therefore, unlike Sports-1M and ActivityNet, YouTube-8M is not
restricted to action classes alone. For example, Figure~\ref{fig:examples} shows random video examples for the {\it Guitar} entity.

We first construct a \emph{visual annotation vocabulary} from Knowledge Graph entities that appear as topic annotations for
YouTube videos based on the YouTube video annotation system~\cite{youtube-legos}.
To ensure that our vocabulary consists of entities that are recognizable visually,
we use various filtering criteria, including human raters. The entities in the dataset span activities (sports, games,
hobbies), objects (autos, food, products), scenes (travel), and events. The entities were selected using a combination of their popularity on YouTube and
manual ratings of their {\em visualness} according to human raters.
They are an attempt to describe the central themes of videos using a few succinct labels.

We then collect a sample set of videos for each entity, and use a publicly available state-of-the-art Inception network~\cite{tensorflowimage} to extract features from them.
Specifically, we decode videos at one frame-per-second and extract the last hidden representation
before the classification layer for each frame. We compress the frame-level features and make
them available on our website for download.

\begin{figure}[t]
  \centering
  \includegraphics[width=\linewidth]{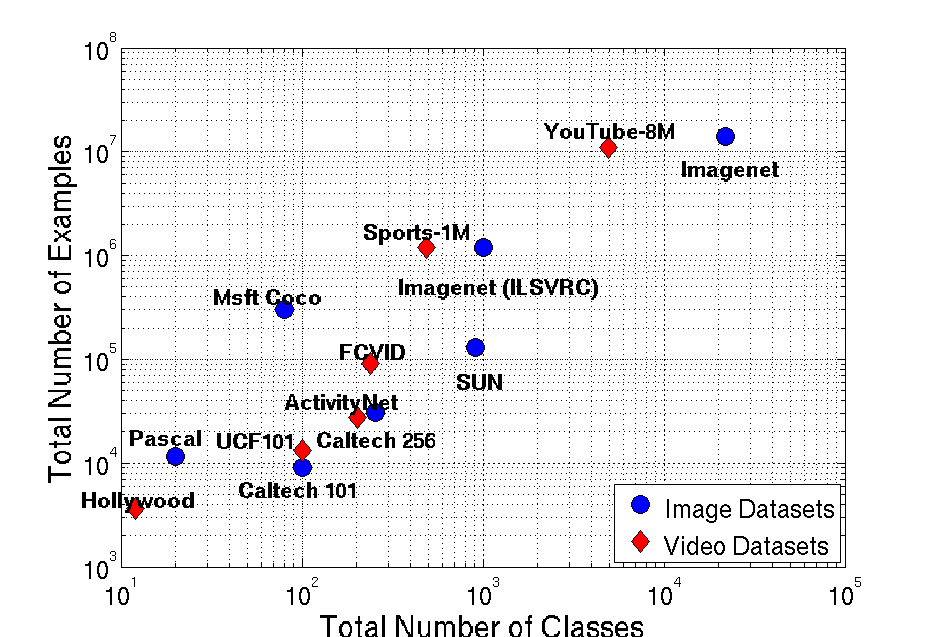}
  \caption{\small The progression of datasets for image and video understanding tasks. Large datasets have played a key role for advances in both areas.}
\label{fig:datasets}
\end{figure}

Overall, YouTube-8M contains more than 8 million videos---over 500,000 hours
of video---from 4,800 classes. Figure~\ref{fig:datasets} illustrates the
scale of YouTube-8M, compared to existing image and video datasets.
We hope that the unprecedented scale and diversity of this dataset will be a
useful resource for developing advanced video understanding and representation
learning techniques.

Towards this end, we provide extensive experiments comparing several state-of-the-art techniques for video representation learning, including
 Deep Networks~\cite{beyond}, and LSTMs (Long Short-Term Memory Networks)~\cite{hochreiter97long} on this dataset. In addition, we show that transfering video feature representations learned on this dataset leads to significant improvements on other benchmarks such as Sports-1M and ActivityNet.

In the rest of the paper, we first review existing benchmarks for image and video classification in Section~\ref{sec:related}. We present the details of our dataset including the
collection process and a brief analysis of the categories and videos in Section~\ref{sec:dataset}. In Section~\ref{sec:benchmarks}, we review several approaches for the task of multi-label video classification given fixed frame-level features, and evaluate the approaches on the dataset. In Section~\ref{sec:experiments}, we show that features and models learned on our large-scale dataset generalize very well on other benchmarks. We offer concluding remarks with Section~\ref{sec:conclusions}.

\section{Related Work}
\label{sec:related}

Image benchmarks have played a significant role in advancing computer vision
algorithms for image understanding. Starting from a number of well labeled
small-scale datasets such as
Caltech 101/256\hfill~\cite{caltech-101,caltech-256},
MSRC~\cite{msrc}, PASCAL~\cite{pascal}, image understanding research has rapidly
advanced to utilizing larger datasets such as ImageNet~\cite{ImageNet} and
SUN~\cite{sun} for the next generation of vision algorithms. ImageNet in
particular has enabled the development of deep feature learning techniques with
millions of parameters such as the AlexNet\hfill~\cite{krizhevsky2012imagenet} and
Inception~\cite{inception-bn} architectures due to the number of classes ($21841$),
the diversity of the classes ($27$ top-level categories) and the millions of
labeled images available.

A similar effort is in progress in the video understanding domain where the community has quickly progressed from small, well-labeled datasets such as
 KTH~\cite{kth}, Hollywood 2~\cite{hollywood}, Weizmann~\cite{weizmann}, with a few thousand video clips, to medium-scale datasets such as UCF101~\cite{ucf101},
 Thumos`14~\cite{thumos} and HMDB51~\cite{hmdb}, with more than 50 action categories. Currently, the largest available video benchmarks are the
 Sports-1M~\cite{karpathy2014large}, with $487$ sports related activities and $1$M videos, the YFCC-100M~\cite{yfcc}, with $800$K videos and
 raw metadata (titles, descriptions, tags) for some of them, the FCVID~\cite{fcvid} dataset of $91,223$ videos manually annotated with $239$ categories, and
 ActivityNet~\cite{actnet}, with $\mathtt{\sim}200$ human activity classes and a few thousand videos. However,
almost all current video benchmarks are restricted to recognizing action and activity categories, and have less than 500 categories.

\begin{figure*}[t]
\begin{center}
  \includegraphics[width=0.95\linewidth]{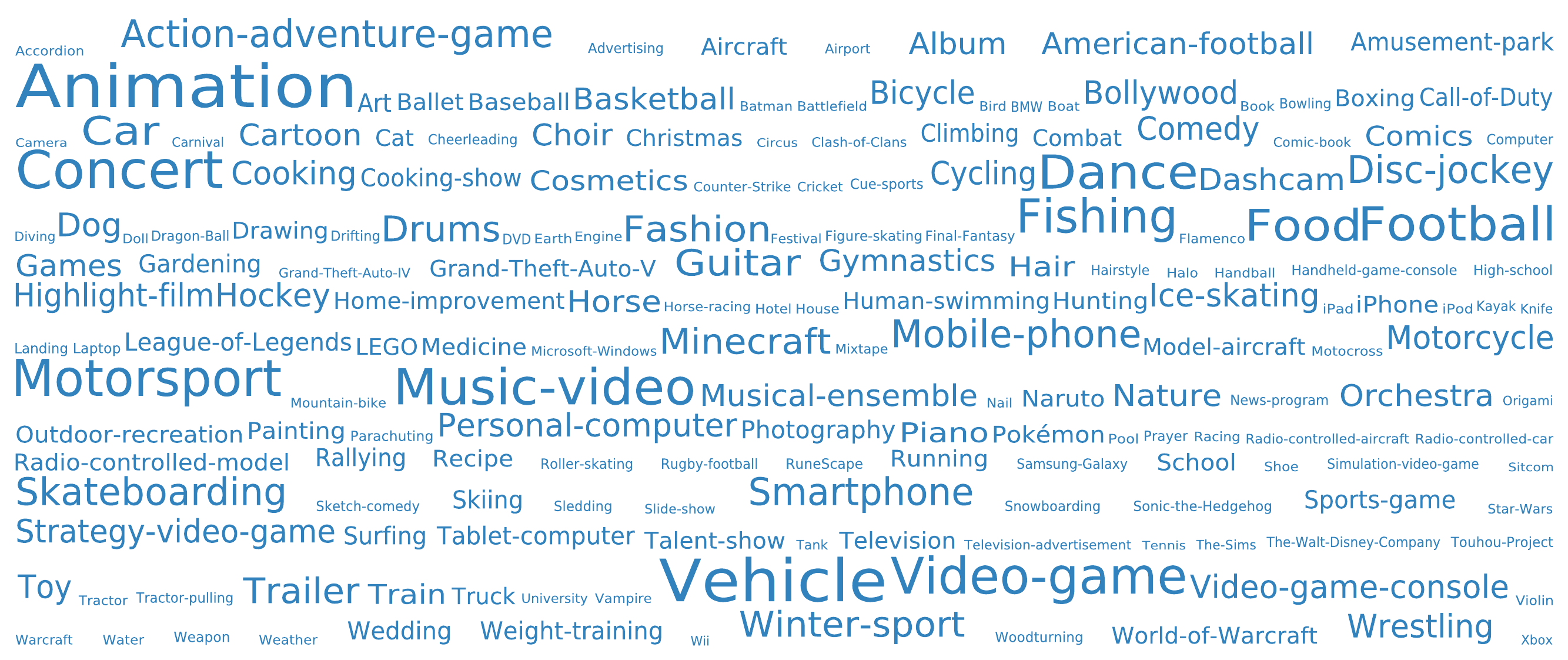}
\end{center}
\caption{A tag-cloud representation of the top 200 entities. Font size is
  proportional to the number of videos labeled with the entity.}
\label{fig:tag-cloud}
\end{figure*}

\begin{table*}
\centering \small
\scalebox{0.8}{
\begin{tabular}{|l|l|l|l|l|l|l|l|l|l|l|}
\hline %
\bf{Top-level Category} & 1\textsuperscript{st} Entity & 2\textsuperscript{nd} Entity & 3\textsuperscript{rd} Entity & 4\textsuperscript{th} Entity & 5\textsuperscript{th} Entity & 6\textsuperscript{th} Entity & 7\textsuperscript{th} Entity \\ \hline%
\bf{Arts \& Entertainment} & Concert & Animation & Music video & Dance & Guitar & Disc jockey & Trailer \\
\bf{Autos \& Vehicles} & Vehicle & Car & Motorcycle & Bicycle & Aircraft & Truck & Boat \\
\bf{Beauty \& Fitness} & Fashion & Hair & Cosmetics & Weight training & Hairstyle & Nail & Mascara \\
\bf{Books \& Literature} & Book & Harry Potter & The Bible & Writing & Magazine & Alice & E-book \\
\bf{Business \& Industrial} & Train & Model aircraft & Fish & Water & Tractor pulling & Advertising & Landing \\
\bf{Computers \& Electronics} & Personal computer & Video game console & iPhone & PlayStation 3 & Tablet computer & Xbox 360 & Microsoft Windows \\
\bf{Finance} & Money & Bank & Foreign Exchange & Euro & United States Dollar & Credit card & Cash \\
\bf{Food \& Drink} & Food & Cooking & Recipe & Cake & Chocolate & Egg & Eating \\
\bf{Games} & Video game & Minecraft & Action-adventure game & Strategy video game & Sports game & Call of Duty & Grand Theft Auto V \\
\bf{Health} & Medicine & Raw food & Ear & Glasses & Injury & Dietary supplement & Dental braces \\
\bf{Hobbies \& Leisure} & Fishing & Outdoor recreation & Radio-controlled model & Wedding & Christmas & Hunting & Diving \\
\bf{Home \& Garden} & Gardening & Home improvement & House & Kitchen & Garden & Door & Swimming pool \\
\bf{Internet \& Telecom} & Mobile phone & Smartphone & Telephone & Website & Sony Xperia & Google Nexus & World Wide Web \\
\bf{Jobs \& Education} & School & University & High school & Teacher & Kindergarten & Campus & Classroom \\
\bf{Law \& Government} & Tank & Firefighter & President of the U.S.A. & Soldier & President & Police officer & Fighter aircraft \\
\bf{News} & Weather & Snow & Rain & News broadcasting & Newspaper & Mattel & Hail \\
\bf{People} \& Society & Prayer & Family & Play-Doh & Human & Dragon & Angel & Tarot \\
\bf{Pets} \& Animals & Animal & Dog & Horse & Cat & Bird & Aquarium & Puppy \\
\bf{Real Estate} & House & Apartment & Condominium & Dormitory & Mansion & Skyscraper & Loft \\
\bf{Reference} & Vampire & Bus & River & City & Mermaid & Village & Samurai \\
\bf{Science} & Nature & Robot & Eye & Ice & Biology & Skin & Light \\
\bf{Shopping} & Toy & LEGO & Sledding & Doll & Shoe & My Little Pony & Nike; Inc. \\
\bf{Sports} & Motorsport & Football & Winter sport & Cycling & Basketball & Gymnastics & Wrestling \\
\bf{Travel} & Amusement park & Hotel & Airport & Beach & Roller coaster & Lake & Resort \\\hline%
\bf{Full vocabulary} & \bf{Vehicle} & \bf{Concert} & \bf{Animation} &\bf{ Music video} &\bf{ Video game} & \bf{Motorsport} & \bf{Football} \\
\hline
\end{tabular}
}
\caption{Most frequent entities for each of the top-level categories.}
\label{table:verticals-top-entities}
\end{table*}

YouTube-8M fills the gap in video benchmarks as follows:
\begin{itemize} \setlength\itemsep{0em}
  \item A \textbf{large-scale video annotation and representation learning benchmark}, reflecting the main themes of a video.%
  \item A significant jump in the number and diversity of annotation classes---\textbf{4800 Knowledge Graph entities} vs. less than 500 categories for all other datasets.
  \item A substantial increase in the number of labeled videos---over \textbf{8~million videos, more than 500,000~hours of video}.
  \item Availability of \textbf{pre-computed state-of-the-art features for 1.9~billion video frames}.
\end{itemize}
 We hope the pre-computed features will remove computational barriers, level the playing field, and enable researchers to explore new technologies in the video domain at an unprecedented scale.

\section{YouTube-8M Dataset}
\label{sec:dataset}

\begin{figure*}
  \begin{subfigure}[ht]{0.53\linewidth}
    \includegraphics[width=\linewidth]{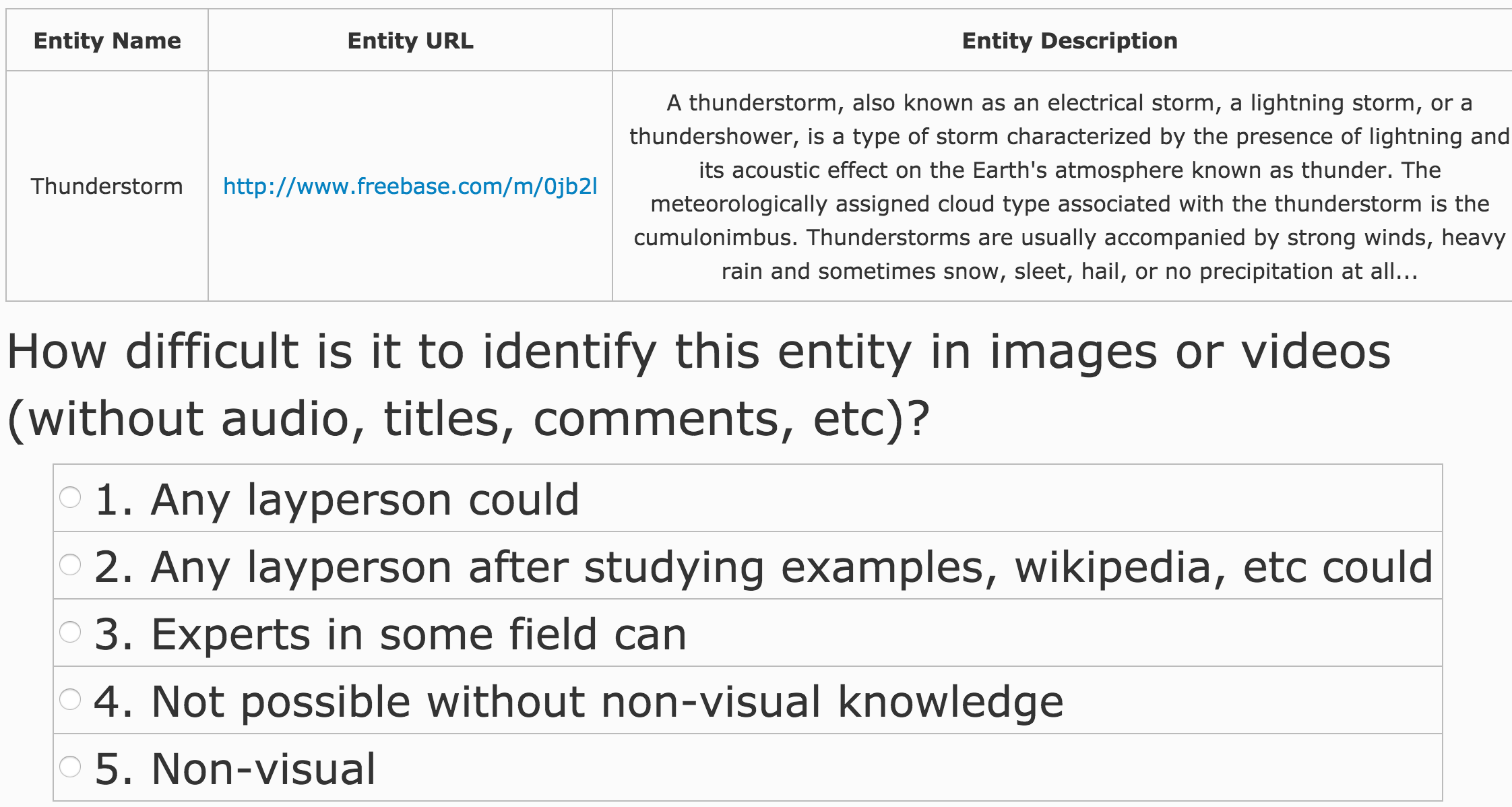}
    \caption{\label{fig:mturkraters} Screenshot of the question displayed to human raters.}
  \end{subfigure}
  \hspace{0.2in}
  \begin{subfigure}[ht]{0.45\linewidth}
    \centering
    \includegraphics[width=\linewidth]{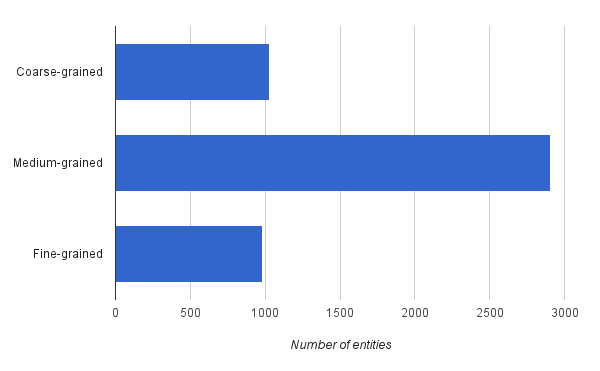}
    \caption{\label{fig:visualness} Distribution of vocabulary topics in terms of specificity.}
  \end{subfigure}
  \caption{Rater guidelines to assess how specific and visually recognizable each entity is,
    on a discrete scale of (1 to 5), where 1 is most visual and easily recognizable by a layperson.
    Each entity was rated by 3 raters. We kept only entities with a maximum average
    score of $2.5$, and categorized them by specificity, into coarse-grained,
    medium-grained, and fine-grained entities, using equally sized score range buckets.}
\end{figure*}

YouTube-8M is a benchmark dataset for video understanding, where the
main task is to determine the key topical themes of a video. We start with
YouTube videos since they are a good (albeit noisy) source of knowledge for
diverse categories including various sports, activities, animals, foods, products,
tourist attractions, games, and many more. We use the YouTube video annotation system~\cite{youtube-legos}
to obtain topic annotations for a video, and to retrieve videos for a given
topic. The annotations are provided in the form of Knowledge Graph entities~\cite{kg} (formerly, Freebase topics~\cite{freebase}).
They are associated with each video based on the video's metadata, context, and content signals~\cite{youtube-legos}.

We use Knowledge Graph entities to succinctly describe the {\em main themes} of
a video. For example, a video of biking on dirt roads
and cliffs would have a central topic/theme of {\it Mountain Biking}, not {\it Dirt},
{\it Road}, {\it Person}, {\it Sky}, and so on. Therefore, the aim of the dataset is not
only to understand what is present in each frame of the video, but also to identify
the few key topics that best describe what the video is about. Note that this is different
than typical event or scene recognition tasks, where each item belongs to a
single event or scene.~\cite{sun,indoorscene}  It is also different than
most object recognition tasks, where the goal is to label everything visible
in an image. This would produce thousands of labels on each video but without
answering what the video is really about. The goal of this benchmark is to
understand what is in the video and to summarize that into a few key topics.
In the following sub-sections, we describe our vocabulary and video
selection scheme, followed by a brief summary of dataset statistics.

\subsection{Vocabulary Construction}

We followed two main tenets when designing the vocabulary for the dataset;
namely 1) every label in the dataset should be distinguishable using visual information
alone, and 2) each label should have sufficient number of videos for training models
and for computing reliable metrics on the test set. For the former, we used a combination of manually
curated topics and human ratings to prune the vocabulary into a visual set. For the latter, we
considered only entities having at least $200$ videos in the dataset.

The Knowledge Graph contains millions of topics. Each topic has one or more {\it types},
that are curated with high precision. For example, there is an exhaustive list of
animals with type {\it animal} and an exhaustive list of foods with type {\it food}.
To start with our initial vocabulary, we manually selected a whitelist of 25
entity types that we considered visual (e.g. {\it sport},
{\it tourist\_attraction}, {\it inventions}), and also blacklisted types that
we thought are non-visual (e.g. {\it music artists}, {\it music compositions}, {\it album}, {\it software}).
We then obtained all entities that have at least one whitelisted type and no
blacklisted types, which resulted in an initial vocabulary of $\mathtt{\sim}50,000$ entities.

Following this, we used human raters in order to manually
prune this set into a smaller set of entities that are considered visual with high
confidence, and are also recognizable without very deep domain expertise. Raters were provided with instructions and examples. Each entity was
rated by 3 raters and the ratings were averaged. Figure \ref{fig:mturkraters} shows the main rating question.
The process resulted in a total of $\mathtt{\sim}10,000$ entities that are considered visually recognizable
and are not too fine-grained (i.e. can be recognized by non-domain experts after studying some examples).
These entities were further pruned: we only kept entities that have more than 200 popular videos,
as explained in the next section.  The final set of entities in the dataset are
fairly balanced in terms of the specificity of the topic they describe, and span
both coarse-grained and fine-grained entities, as shown in Figure~\ref{fig:visualness}.

\subsection{Collecting Videos}

Having established the initial target vocabulary, we followed these steps to obtain
the videos:

\begin{itemize}
  \setlength\itemsep{0em}
  \item Collected all videos corresponding to the $~10,000$ visual entities and have at least $1,000$ views, using the YouTube video annotation system~\cite{youtube-legos}. We excluded too short (< 120 secs) or too long (> 500 secs) videos. %
  \item Randomly sampled 10 million videos among them.
  \item Obtained all entities for the sampled 10 million videos using the YouTube video annotation system. This completes the annotations.
  \item Filtered out entities with less than $200$ videos, and videos with no remaining entities. This reduced the size of our data to $8,264,650$ videos. %
  \item Split our videos into 3 partitions, {\it Train} : {\it Validate} : {\it Test}, with ratios $70\% : 20\% : 10\%$. We publish features for all splits, but only publish labels for the {\it Train} and {\it Validate} partitions.
\end{itemize}

\subsection{Features}
\label{sec:features}

The original size of the video dataset is hundreds of Terabytes, and covers over $500,000$ hours of video.
This is impractical to process by most research teams (using a real-time video
processing engine, it would take over 50 years to go through the data). Therefore, we
pre-process the videos and extract frame-level features using a state-of-the-art deep model:
the publicly available Inception network~\cite{tensorflowimage} trained on ImageNet~\cite{inception-bn}.
Concretely, we decode each video at 1 frame-per-second up to the first 360 seconds
(6 minutes), feed the decoded frames into the Inception network, and
fetch the ReLu activation of the last hidden layer, before the classification layer (layer name
\texttt{pool\_3/\_reshape}). The feature vector is 2048-dimensional per second of video. While this removes motion
information from the videos, recent work shows diminishing returns from motion
features as the size and diversity of the video data increases~\cite{beyond,c3d}. The static
frame-level features provide an excellent baseline, and constructing compact and
efficient motion features is beyond the scope of this paper. Nonetheless, we hope to
extend the dataset with audio and motion features in the future.
We cap processing of each video up to the first
$360$ seconds for storage and computational reasons. For comparison, the average length of videos in UCF-101 is
$10-15$ seconds, Sports-1M is $336$ seconds and in this dataset, it is $230$ seconds.

Afterwards, we apply PCA ($+$ whitening) to reduce feature dimensions to 1024, followed
by quantization (1 byte per coefficient). These two compression techniques reduce the size of the data by a
factor of 8. The mean vector and covariance matrix for PCA was computed on all
frames from the {\it Train} partition.
We quantize each 32-bit float into 256 distinct values (8 bits) using optimally computed (non-uniform) quantization bin boundaries.
We confirmed that the size reduction does not significantly hurt the evaluation metrics. In fact,
training all baselines on the full-size data (8 times larger than what we publish),
increases all evaluation metrics by less than $1\%$.

\begin{table}
  \begin{tabular}{| l | r | r | r | r |}
    \hline
    Dataset & Train & Validate & Test & Total \\
    \hline \hline
    YouTube-8M & 5,786,881 & 1,652,167 & 825,602 & 8,264,650 \\
    \hline
  \end{tabular}
\caption{Dataset partition sizes.}
\label{table:numvideos}
\end{table}

Note that while this dataset comes with standard frame-level features, it leaves
a lot of room for investigating video representation learning approaches on top
of the fixed frame-level features (see Section~\ref{sec:benchmarks} for approaches we explored).

\subsection{Dataset Statistics}
\label{sec:stats}

\begin{figure}
\begin{center}
  \includegraphics[width=\linewidth]{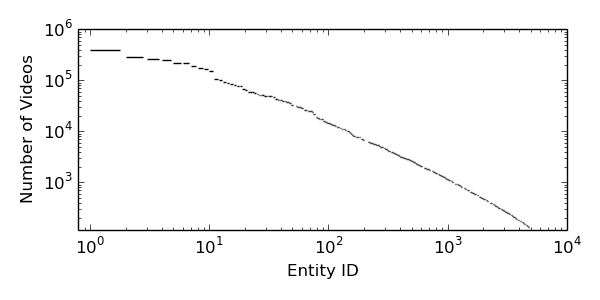}
\caption{Number of videos in log-scale versus entity rank in log scale.
         Entities were sorted by number of videos. We note that this somewhat
         follows the natural Zipf distribution.}
\label{fig:entityvideos}
\end{center}
\end{figure}

\begin{figure*}
\begin{center}
  \begin{subfigure}[ht]{0.48\linewidth}
  \begin{center}
    \includegraphics[height=180pt]{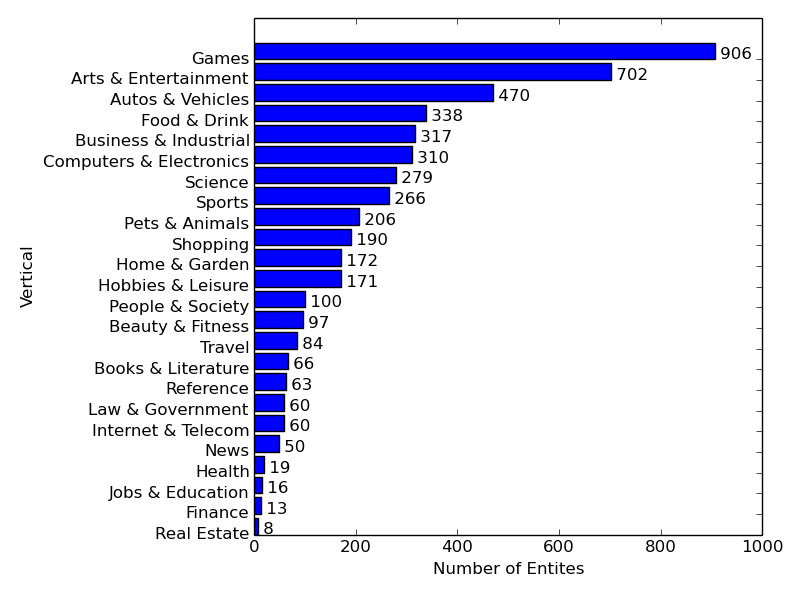}
  \end{center}
  \caption{Number of entities in each top-level category.}
  \label{fig:verticalentities}
  \end{subfigure}
  \begin{subfigure}[ht]{0.48\linewidth}
  \begin{center}
    \includegraphics[height=180pt]{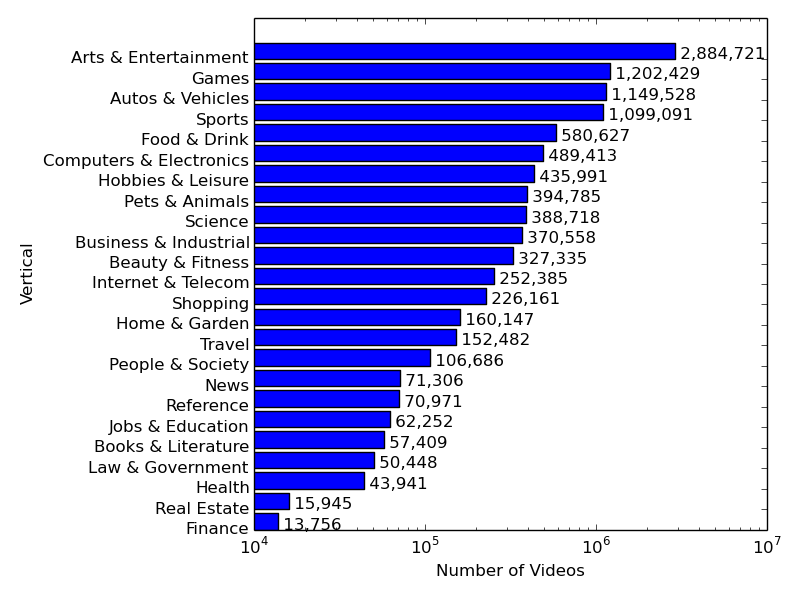}
  \caption{Number of {\it train} videos in log-scale per top-level category.}
  \label{fig:verticalvideos}
  \end{center}
  \end{subfigure}
\end{center}
\vspace{-0.2in}
\caption{Top-level category statistics of the YouTube-8M dataset.}
\end{figure*}

The YouTube-8M dataset contains $4,800$ classes and a total of
$8,264,650$ videos. A video may be annotated with more than one class and
the average number of classes per video is $1.8$. Table~\ref{table:numvideos} shows the number of videos for which we are releasing features,
across the three datasets.

We processed only the first six minutes of each video, at 1 frame-per-second.
The average length of a video in the dataset is
229.6 seconds, which amounts to $\mathtt{\sim}1.9$
billion frames (and corresponding features) across the dataset.

We grouped the $4,800$ entities into 24 top-level categories to measure statistics and illustrate diversity. Although
we do not use these categories during training, we are releasing the entity-to-category
mapping for completeness. Table~\ref{table:verticals-top-entities} shows
the top entities per category. Note that while some categories themselves may not seem
visual, most of the entities within them are visual. For instance, Jobs \& Education includes universities, classrooms, lectures, etc., and
Law \& Government includes police, emergency vehicles, military-related entities, which are well represented and visual.

Figure~\ref{fig:entityvideos} shows a log-log scale distribution
of entities and videos. Figures~\ref{fig:verticalentities} and \ref{fig:verticalvideos}
show the size of categories, respectively, in terms of the number of entities and
the number of videos.

\subsection{Human Rated Test Set}
\label{sec:humangt}
The annotations from the YouTube video annotation system can be noisy and incomplete, as they are automatically
generated from metadata, anchor text, comments, and user engagement signals~\cite{youtube-legos}.
To quantify the noise, we uniformly sampled over $8000$ videos from
the {\it Test} partition, and used 3 human raters per video to exhaustively rate their labels.
We measured the precision and recall of the ground truth labels to be $78.8\%$ and $14.5\%$,
respectively, with respect to the human raters. Note that typical inter-rater agreement
on similar annotation tasks with human raters is also around 80\% so the precision of
these ground truth labels is perhaps comparable to (non-expert) human-provided labels.
The recall, however, is low, which makes this an excellent test bed for approaches
that deal with missing data.  We report the accuracy of our models primarily on the
(noisy) {\it Validate} partition but also show some results on the much smaller human-rated set,
showing that some of the metrics are surprisingly similar on the two datasets.

While the baselines in section~\ref{sec:benchmarks} show very promising results, we believe
that they can be significantly improved (when evaluated on the human-based ground truth),
if one explicitly models incorrect~\cite{reed2014noisy} ($78.8\%$ precision)
or missing~\cite{missinglabels, noisydata} ($14.5\%$ recall) training labels.
We believe this is an exciting area of research that this dataset will enable at scale.

\section{Baseline Approaches}
\label{sec:benchmarks}

\subsection{Models from Frame Features}
\label{sec:frame_level_models}
One of the challenges with this dataset is that we only have video-level
ground-truth labels. We do not have any additional information that specifies how
the labels are localized within the video, nor their relative prominence in the
video, yet we want to infer their importance for the full video. In this section, we
consider models trained to predict the main themes of the video using the input
frame-level features. Frame-level models have shown competitive performance for
video-level tasks in previous work~\cite{karpathy2014large,beyond}. A
video $v$ is given by a sequence of frame-level features $\mathbf{x}^v_{1:F_v}$,
where $\mathbf{x}^v_j$ is the feature of the $j^{th}$ frame from video $v$.

\subsubsection{Frame-Level Models and Average Pooling}
\label{sec:frame_level_features}
Since we do not have frame-level ground-truth, we assign the video-level
ground-truth to every frame within that video. More sophisticated formulations
based on multiple-instance learning are left for future work. From each video,
we sample 20 random frames and associate all frames to the video-level
ground-truth. This results in about $120$ million frames. For each entity $e$,
we get $120M$ instances of $(\mathbf{x}_i, y^{e}_i)$ pairs, where
$\mathbf{x}_i \in \mathbb{R}^{1024}$ is the inception feature and $y^e_i \in {0,1}$
is the ground-truth associated with entity $e$ for the $i^{th}$ example. We train
$4800$ independent one-vs-all classifiers for each entity $e$. We use the online
training framework after parallelizing the work for each entity across multiple
workers. During inference, we score every frame in the test video using the models
for all classes. Since all our evaluations are based on video-level ground truths,
we need to aggregate the frame-level scores (for each entity) to a single
video-level score. The frame-level probabilities are aggregated to the video-level
using a simple average. We choose average instead of max pooling since we want to
reduce the effect of outlier detections and capture the prominence of each entity
in the entire video. In other words, let $p(e | \mathbf{x})$ be the probability
of existence of $e$ given the features $\mathbf{x}$. We compute the probability
$p_v(e | \mathbf{x}^v_{1:F_v})$ of the entity $e$ associated with the video $v$ as

\begin{equation}
p_v(e | \mathbf{x}^v_{1:F_v}) = \frac{1}{F_v} \sum_{j=1}^{F_v} p(e | \mathbf{x}^v_j).
\end{equation}

\subsubsection{Deep Bag of Frame (DBoF) Pooling}
\label{sec:dbof}
\begin{figure}
\begin{center}
  \includegraphics[width=0.5\linewidth]{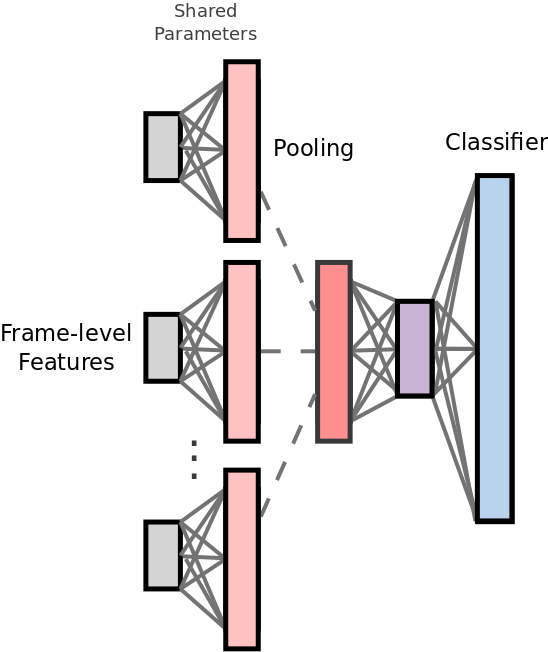}
  \caption{\small The network architecture of the DBoF approach. Input frame
  features are first fed into a up-projection layer with shared parameters for
  all frames. This is followed by a pooling layer that converts the frame-level
  sparse codes into a video-level representation. A few hidden layers and a
  classification layer provide the final video-level predictions.}
\label{fig:mpool}
\end{center}
\end{figure}

Inspired by the success of various classic bag of words representations for video classification~\cite{hollywood, Wang09evaluationof}, we next consider a Deep
 Bag-of-Frames (DBoF) approach. Figure~\ref{fig:mpool} shows the overall architecture of our DBoF network for video classification. The $N$-dimensional
input frame level features from $k$ randomly selected frames of a video are first fed into a fully connected layer of $M$ units with RELU activations. Typically, with $M > N$, the input features are projected onto a higher dimensional space. Crucially, the parameters of the fully connected layer are shared
across the $k$ input frames. Along with the RELU activation, this leads to a sparse coding of the input features in the $M$-dimensional space.

The obtained sparse codes are fed into a pooling layer that aggregates the codes
of the $k$ frames into a single fixed-length video representation. We use max
pooling to perform the aggregation. We use a batch normalization layer before
pooling to improve stability and speed-up convergence. The obtained fixed length
descriptor of the video can now be classified into the output classes using a
Logistic or Softmax layer with additional fully connected layers in between. The
$M$-dimensions of the projection layer could be thought of as $M$ discriminative
clusters which can be trained in a single network end to end using backpropagation.

The entire network is trained using Stocastic Gradient Descent (SGD) with logistic loss for a logistic layer and cross-entropy loss for a softmax layer. The
backpropagated gradients from the top layer train the weight vectors of the projection layer in a discriminative fashion in order to provide a powerful
representation of the input bag of features. A similar network was proposed in~\cite{beyond} where the convolutional layer outputs are pooled across all the
frames of a video to obtain a fixed length descriptor. However, the network in~\cite{beyond} does not use an intermediate projection layer which we found
to be a crucial difference when learning from input frame features.
Note that the up-projection layer into sparse codes is similar to what Fisher
Vectors~\cite{fisher1} and VLAD~\cite{vlad} approaches do but the projection
(i.e., clustering) is done discriminatively here. We also experimented with
Fisher Vectors and VLAD but were not able to obtain competitive results using
comparable codebook sizes.

\textbf{Hyperparameters:} We considered values of $\{2048, 4096, 8192\}$ for the
number of units in the projection layer of the network and found that larger
values lead to better results. We used $8192$ for all datasets. We used a single
hidden layer with $1024$ units between the pooling layer and the final classification
layer in all experiments. The network was trained using SGD with AdaGrad, a
learning rate of $0.1$, and a weight decay penalty of $0.0005$.

\subsubsection{Long Short-Term Memory (LSTM)}
\label{sec:lstm}
We take a similar approach to~\cite{beyond} to utilize LSTMs for video-level prediction. However, unlike that work, we do not have access to the raw video frames. This means that we can only train the LSTM and Softmax layers.

  We experimented with the number of stacked LSTM layers and the number of hidden units. We empirically found that 2 layers with 1024 units provided the highest performance on the validation set. Similarly to~\cite{beyond}, we also employ linearly increasing per-frame weights going from $1/N$ to $1$ for the last frame.

  During the training time, the LSTM was unrolled for 60 iterations. Therefore, the gradient horizon for LSTM was 60 seconds. We experimented with a larger number of unroll iterations, but that slowed down the training process considerably. In the end, the best model was the one trained for the largest number of steps (rather than the most real time).

  In order to transfer the learned model to ActivityNet, we used a fully-connected
  model which uses as inputs the concatenation of the LSTM layers' outputs as
  computed at the last frame of the videos in each of these two benchmarks.
  Unlike traditional transfer learning methods, we do not fine-tune the LSTM
  layers. This approach is more robust to overfitting than traditional methods,
  which is crucial for obtaining competitive performance on ActivityNet due to
  its size. We did perform full fine-tuning experiments on Sports-1M, which is
  large enough to fine-tune the entire LSTM model after pre-training.

\subsection{Video level representations}
Instead of training classifiers directly on frame-level features, we also
explore extracting a task-independent fixed-length video-level feature vector
from the frame-level features $\mathbf{x}^v_{1:F_v}$ for each video $v$. There
are several benefits of extracting fixed-length video features:
\begin{enumerate} \setlength\itemsep{0em}
\item \textbf{Standard classifiers can apply:} Since the dimensionality of the representations are fixed across videos, we may train standard classifiers like logistic, SVM, mixture of experts.
\item \textbf{Compactness:} We get a compact representation for the entire video, thereby reducing the training data size by a few orders of magnitude.
\item \textbf{More suitable for domain adaptation:} Since the video-level representations are unsupervised (extracted independently of the labels), these representations are far less specialized to the labels associated with the current dataset,
and can generalize better to new tasks or video domains.
\end{enumerate}
Formally, a video-level feature $\varphi(\mathbf{x}^v_{1:F_v})$ is a fixed-length representation (at the video-level). We explore a simple aggregation technique for getting these video-level representations.
We also experimented with Fisher Vectors~(FV)~\cite{fisher1} and VLAD~\cite{vlad}
approaches for task-independent video-level representations but were not able to
achieve competitive results for FV or VLAD representations of similar dimensionality.
We leave it as future work to come up with compact FV or VLAD type representations
that outperform the much simpler approach described below.

\subsubsection{First, second order and ordinal statistics}
\label{sec:simplestats}
From the frame-level features $\mathbf{x}^v_{1:F_v}$ where $\mathbf{x}^v_j \in \mathbb{R}^{1024}$,
we extract the mean $\mathbf{\mu}^v \in \mathbb{R}^{1024}$ and the standard-deviation
$\mathbf{\sigma}^v \in \mathbb{R}^{1024}$. Additionally, we also extract the top $5$
ordinal statistics for each dimension. Formally, $\operatorname{Top}_K(\mathbf{x}^v(j)_{1:F_v})$ returns a $K$ dimensional vector where the $p^{th}$ dimension contains the $p^{th}$ highest value of the feature-vector's $j^{th}$ dimension over the entire video. We denote $\operatorname{Top}_K(\mathbf{x}^v_{1:F_v})$ to be a $KD$ dimensional vector obtained by concatenating the ordinal statistics for each dimension. Thus, the resulting feature-vector  $\varphi(\mathbf{x}^v_{1:F_v})$ for the video becomes:
\begin{equation}
\varphi(\mathbf{x}^v_{1:F_v}) = \begin{bmatrix}
\mathbf{\mu}(\mathbf{x}^v_{1:F_v}) \\
\mathbf{\sigma}(\mathbf{x}^v_{1:F_v}) \\
\operatorname{Top}_K(\mathbf{x}^v_{1:F_v})
\end{bmatrix}.
\label{eq:videofeature}
\end{equation}

\subsubsection{Feature normalization}
\label{sec:normalization}
Standardization of features has been proven to help with online learning
algorithms\cite{inception-bn,mean_normalized_sgd} as it makes the updates using
Stochastic Gradient Descent (SGD) based algorithms (like Adagrad) more robust to
learning rates, and speeds up convergence.

Before training our one-vs-all classifiers on the video-level representation, we
apply global normalization to the feature vectors $\varphi(\mathbf{x}^v_{1:F_v})$
(defined in equation \ref{eq:videofeature}). Similar to how we processed the frame features,
we substract the mean $\varphi(.)$ then use PCA to decorrelate and whiten the features.
The normalized video features are now approximately multivariate gaussian with zero mean
and identity covariance. This makes the gradient steps across the various dimensions
independent, and learning algorithm gets an unbiased view of each dimension (since
the same learning rate is applied to each dimension). Finally, the resulting
features are $L_2$ normalized. We found that these normalization techniques
make our models train faster.

\subsection{Models from Video Features} \label{sec:binarymodels}
Given the video-level representations, we train independent binary classifiers
for each label using all the data. Exploiting the structure information between
the various labels is left for future work. A key challenge is training
these classifiers at the scale of this dataset. Even with a compact video-level
representation for the $6M$ training videos, it is unfeasible to train batch
optimization classifiers, like SVM. Instead, we use online learning algorithms,
and use Adagrad to perform model updates on the weight vectors given a small
mini-batch of examples (each example is associated with a binary ground-truth value).

\subsubsection{Logistic Regression}
Given $D$ dimensional video-level features, the parameters $\mathbf{\Theta}$ of the logistic regression classifier are the entity specific weights $\mathbf{w}_e$. During scoring, given $\mathbf{x} \in \mathbb{R}^{D+1}$ to be the video-level feature of the test example, the probability of the entity $e$ is given as $p(e | \mathbf{x}) = \sigma(\mathbf{w}_e^T \mathbf{x})$. The weights $\mathbf{w}_e$ are obtained by minimizing the total log-loss on the training data given as:
\begin{equation}
\lambda \Vert \mathbf{w}_e \Vert_{2}^2 + \sum_{i=1}^N \mathcal{L}(y_{i,e}, \sigma(\mathbf{w}_e^T \mathbf{x}_i)),
\end{equation}

where $\sigma(.)$ is the standard logistic, $\sigma(z) = 1 / (1 + \exp(-z))$.

\subsubsection{Hinge Loss}
Since training batch SVMs on such a large dataset is impossible, we use the online SVM approach. As in the conventional SVM framework, we use $\pm 1$ to represent negative and positive labels respectively. Given binary ground-truth labels $y$ ($0$ or $1$), and predicted labels $\hat{y}$ (positive or negative scalars), the hinge loss is:
\begin{equation}
\mathcal{L}(y, \hat{y}) = \operatorname{max}(0, b-(2y-1)\hat{y}),
\end{equation}
where $b$ is the hinge-loss parameter which can be fine-tuned further or set to $1.0$. Due to the presence of the $\operatorname{max}$ function, there is a discontinuity in the first derivative. This results in the subgradient being used in the updates, slowing convergence significantly.

\subsubsection{Mixture of Experts (MoE)}
Mixture of experts (MoE) was first proposed by Jacobs and Jordan~\cite{Jordan94hierarchicalmixtures}. The binary classifier for an entity $e$ is composed of a set of hidden states, or experts, $\mathcal{H}_e$. A softmax is typically used to model the probability of choosing each expert. Given an expert, we can use a sigmoid to model the existence of the entity. Thus, the final probability for entity $e$'s existence is $p(e | \mathbf{x}) = \sum_{h \in \mathcal{H}_e} p(h | \mathbf{x}) \sigma(\mathbf{u}_{h}^T \mathbf{x})$, where  $p(h | \mathbf{x})$ is a softmax over $|\mathcal{H}_e|+1$ states. In other words, $p(h | \mathbf{x}) = \frac{\exp(\mathbf{w}_h^{T} \mathbf{x})}{1 + \sum_{h' \in \mathcal{H}_e} \exp(\mathbf{w}_{h'}^{T} \mathbf{x})}$.
The last, $(|\mathcal{H}_e|+1)^{th}$, state is a dummy state that always results in the non-existence of the entity. Denote $p_{y|\mathbf{x}} = p(y=1 | \mathbf{x})$, $p_{h|\mathbf{x}} = p(h | \mathbf{x})$ and $p_{h} = p(y=1 | \mathbf{x},h)$. Given a set of training examples ${(\mathbf{x}_i, g_i)}_{i=1\ldots N}$ for a binary classifier, where $\mathbf{x}_i$ is the feature vector and $g_i \in [0,1]$ is the ground-truth, let $\mathcal{L}(p_i,g_i)$ be the log-loss between the predicted probability and the ground-truth:
\begin{equation}
\mathcal{L}(p,g) = -g \log p - (1-g) \log(1-p).
\end{equation}
We could directly write the derivative of $\mathcal{L} \left[p_{y|\mathbf{x}}, g\right]$ with respect to the softmax weight $\mathbf{w}_h$ and the logistic weight $\mathbf{u}_h$ as
\begin{eqnarray}
\frac{\partial \mathcal{L} \left[p_{y|\mathbf{x}}, g\right]}{\partial \mathbf{w}_h} &=& \mathbf{x} \frac{p_{h|\mathbf{x}} \left(p_{y|h,\mathbf{x}} - p_{y|\mathbf{x}}\right) \left(p_{y|\mathbf{x}}-g\right)}{p_{y|\mathbf{x}}(1-p_{y|\mathbf{x}})}, \\
\frac{\partial \mathcal{L} \left[p_{y|\mathbf{x}}, g\right]}{\partial \mathbf{u}_h} &=& \mathbf{x} \frac{p_{h|\mathbf{x}} p_{y|h,\mathbf{x}} (1- p_{y|h,\mathbf{x}})\left(p_{y|\mathbf{x}}-g\right)}{p_{y|\mathbf{x}}(1-p_{y|\mathbf{x}})}.
\end{eqnarray}
We use Adagrad with a learning rate of $1.0$ and batch size of $32$ to learn the weights. Since we are training independent classifiers for each label, the work is distributed across multiple machines.

For MoE models, we experimented with varying number of mixtures ($1$, $2$, $4$),
and found that performance increases by 0.5\%-1\% on all metrics as we go from
$1$ to $2$, and then to $4$ mixtures, but the number of model parameters
correspondingly increases by $2$ or $4$ times. We chose $2$~mixtures as a good
compromise and report numbers with the $2$-mixture MoE model for all datasets.

\section{Experiments}
\label{sec:experiments}
In this section, we first provide benchmark baseline results for the above multi-label classification approaches on the YouTube-8M dataset.
We then evaluate the usefulness of video representations learned on this dataset for other tasks, such as Sports-1M sports classification and AcitvityNet activity classification.

\subsection{Evaluation Metrics}
\textbf{Mean Average Precision (mAP):} For each entity, we first round the annotation scores in buckets of $10^{-4}$  and sort all the \textit{non-zero} annotations according to the model score. At a given threshold $\tau$, the precision $P(\tau)$ and recall $R(\tau)$ are given by
\begin{align}
P(\tau) &= \frac{\sum_{t \in T} \mathbb{I}(\mathbf{y}_t \geq \tau) g_t}{\sum_{t \in T}\mathbb{I}(\mathbf{y}_t \geq \tau)},\\
R(\tau) &= \frac{\sum_{t \in T} \mathbb{I}(\mathbf{y}_t \geq \tau) g_t}{\sum_{t \in T} g_t},
\end{align}

where $\mathbb{I}(.)$ is the indicator function. The average precision, approximating the area under the precision-recall curve, can then be computed as
\begin{equation}
\operatorname{AP} = \sum_{j=1}^{10000} P(\tau_j) [R(\tau_{j}) - R(\tau_{j+1})],
\end{equation}
where where $\tau_j = \frac{j}{10000}$. The mean average precision is computed as the \textit{unweighted} mean of all the per-class average precisions.

\textbf{Hit@$\textbf{k}$:} This is the fraction of test samples that contain at least one of the ground truth labels in the top~$k$ predictions. If $\operatorname{rank}_{v,e}$ is the rank of entity $e$ on video $v$ (with the best scoring entity having rank 1), and $G_v$ is the set of ground-truth entities for $v$, then Hit@$k$ can be written as:
\begin{equation}
\frac{1}{|V|} \sum_{v \in V} \lor_{e \in G_v} \mathbb{I}(\operatorname{rank}_{v,e} \leq k),
\end{equation}
where $\lor$ is logical OR.

\begin{table}
\centering
\scalebox{0.8}{
\begin{tabular}{|l|l|c|c|c|}
\hline
  Input Features & Modeling Approach & mAP & Hit@1 & PERR \\
  \hline \hline
  Frame-level, $\{\mathbf{x}^v_{1:F_v}\}$ & Logistic + Average (\ref{sec:frame_level_features})   & 11.0 & 50.8 & 42.2 \\
  Frame-level, $\{\mathbf{x}^v_{1:F_v}\}$ & Deep Bag of Frames (\ref{sec:dbof})         & 26.9 & 62.7 & 55.1 \\
  Frame-level, $\{\mathbf{x}^v_{1:F_v}\}$ & LSTM (\ref{sec:lstm})                 & 26.6 & \textbf{64.5} & \textbf{57.3} \\
  \hline \hline
  Video-level, $\mathbf{\mu}$ & Hinge loss (\ref{sec:binarymodels})            & 17.0 & 56.3 & 47.9 \\
  Video-level, $\mathbf{\mu}$ & Logistic Regression (\ref{sec:binarymodels})   & 28.1 & 60.5 & 53.0 \\
  Video-level, $\mathbf{\mu}$ & Mixture-of-2-Experts (\ref{sec:binarymodels})  & 29.6 & 62.3 & 54.9 \\
  Video-level, $\left[\mathbf{\mu}; \mathbf{\sigma}; \operatorname{Top}_5\right]$ & Mixture-of-2-Experts (\ref{sec:binarymodels}) & \textbf{30.0} & 63.3 & 55.8 \\
\hline
\end{tabular}
}
\caption{\small Results of the various benchmark baselines on the YouTube-8M dataset. We find that binary classifiers on simple video-level representations perform substantially better than frame-level approaches. Deep learning methods such as DBoF and LSTMs do not provide a substantial boost over traditional dense feature aggregation methods because the underlying frame-level features are already very strong.}
\label{tab:youtube10m}
\end{table}

\begin{table}
\centering
\scalebox{0.8}{
\begin{tabular}{|l|c|c|c|c}
\hline
  Approach &  Hit@1 & PERR & Hit@5 \\
  \hline \hline
  Deep Bag of Frames (DBoF) (\ref{sec:dbof})          & 68.6 & 29.0 & 83.5 \\
  LSTM (\ref{sec:lstm})                               & 69.1 & \textbf{30.5} & \textbf{84.7} \\
  Mixture-of-2-Experts ($\left[\mathbf{\mu}; \mathbf{\sigma}; \operatorname{Top}_5\right]$) (\ref{sec:binarymodels}) & \textbf{70.1} & 29.1 & \textbf{84.8} \\
\hline
\end{tabular}
}
\caption{\small Results of the three best approaches on the human rated test set of the YouTube-8M dataset. A comparison with the results on the validation set (Table~\ref{tab:youtube10m}) shows that the relative strengths of the different approaches are largely
preserved on both sets.}
\label{tab:humangt}
\end{table}

\textbf{Precision at equal recall rate (PERR):} We measure the video-level annotation precision when we retrieve the same number of entities per video as there are in the ground-truth. With the same notation as for Hit@$k$, PERR can be written as:
\begin{equation*}
\frac{1}{|V : |G_v| > 0|} \sum_{v \in V : |G_v| > 0} \left[\frac{1}{|G_v|} \sum_{e \in G_v} \mathbb{I}(\operatorname{rank}_{v,e} \leq |G_v|)\right].
\end{equation*}

\begin{table*}
\centering
\begin{subfigure}[ht]{0.48\linewidth}
\scalebox{0.8}{
\begin{tabular}{|l|c|c|c|}
\hline
  Approach                                                       & mAP  & Hit@1 & Hit@5 \\
\hline \hline
  Logistic Regression ($\mathbf{\mu}$) (\ref{sec:binarymodels})  & 58.0 & 60.1  & 79.6 \\
  Mixture-of-2-Experts ($\mathbf{\mu}$) (\ref{sec:binarymodels})   & 59.1 & 61.5  & 80.4 \\
  Mixture-of-2-Experts ($\left[\mathbf{\mu}; \mathbf{\sigma}; \operatorname{Top}_5\right]$) (\ref{sec:simplestats}) & 61.3 & 63.2 & 82.6 \\
  LSTM (\ref{sec:lstm})                                          & 66.7 & 64.9  & 85.6 \\
  \ \ {\it +Pretrained on YT-8M } (\ref{sec:lstm})               & 67.6 & 65.7  & 86.2 \\
  \hline \hline
  Hierarchical 3D Convolutions~\cite{karpathy2014large}          &   -  & 61.0  & 80.0 \\
  Stacked 3D Convolutions~\cite{c3d}                             &   -  & 61.0  & 85.0 \\
  LSTM with Optical Flow and Pixels~\cite{beyond}                &   -  & \textbf{73.0} & \textbf{91.0} \\
\hline
\end{tabular}
}
\caption{Sports-1M: Our learned features are competitive on this
dataset beating all but the approach of~\cite{beyond}, which learned directly from the video pixels. Both \cite{beyond} and \cite{c3d} included motion features.}
\label{tab:sports1m}
\end{subfigure}
\hspace{0.2in}
\begin{subfigure}[ht]{0.48\linewidth}
\centering
\scalebox{0.8}{
\begin{tabular}{|l|c|c|c|c|}
\hline
  Approach                                              & mAP  & Hit@1 & Hit@5 \\
  \hline \hline
  Mixture-of-2-Experts ($\mathbf{\mu}$) (\ref{sec:binarymodels})  &   69.1 & 68.7  & 85.4 \\
  \ \ {\it +Pretrained PCA on YT-8M }                           &  74.1 & 72.5  & 89.3 \\
  Mixture-of-2-Experts ($\left[\mathbf{\mu}; \mathbf{\sigma}; \operatorname{Top}_5\right]$) (\ref{sec:simplestats}) &  NO & 74.2 & 72.3 & 89.6 \\
  \ \ {\it +Pretrained PCA on YT-8M }                           &  77.6 & 74.9  & 91.6 \\
  LSTM (\ref{sec:lstm})                                         &  57.9 & 63.4  & 81.0 \\
  \ \ {\it +Pretrained on YT-8M } (\ref{sec:lstm})              &  75.6 & 74.2  & 92.4 \\
  \hline \hline
  Ma, Bargal et al.\cite{activitynetsarah}                      &   53.8 &  -    &  -   \\
  Heilbron et al.\cite{actnet}                                  &   43.0 &  -    &  -   \\
\hline
\end{tabular}}
\caption{ActivityNet: Since the dataset is small, we see a substantial boost in
performance by pre-training on YouTube-8M or using the transfer learnt PCA
versus the one learnt from scratch on ActivityNet.}
\label{tab:actnet}
\end{subfigure}
\caption{\small Results of transferring video representations learned on the YouTube-8M dataset to the (a) Sports-1M and (b) ActivityNet. }
\end{table*}

\subsection{Results on YouTube-8M}

Table~\ref{tab:youtube10m} shows results for all approaches on the YouTube-8M
dataset. Frame-level models (row 1), trained on the strong Inception features
and logistic regression, followed by simple averaging of predictions across all
frames, perform poorly on this dataset. This shows that the video-level
prediction task cannot be reduced to simple frame-level classification.

Aggregating the frame-level \emph{features} at the video-level using simple mean
pooling of frame-level features, followed by a hinge loss or logistic regression
model, provides a non-trivial improvement in video level accuracies over naive
averaging of the frame-level predictions. Further improvements are observed by
using mixture-of-experts models and by adding other statistics, like the
standard deviation and ordinal features, computed over the frame-level features.
Note that the standard deviation and ordinal statistics are more meaningful in
the original RELU activation space so we reconstruct the RELU features from the
PCA-ed and quantized features by inverting the quantization and the PCA using
the provided PCA matrix, computing the collection statistics over the reconstructed
frame-level RELU features, and then re-applying PCA, whitening, and L2
normalization as described in Section~\ref{sec:normalization}. This simple
task-independent feature pooling and normalization strategy yields some of the
most competitive results on this dataset.

Finally, we also evaluate two deep network architectures that have produced state-of-art results on previous benchmarks~\cite{beyond}. The DBoF architecture ignores sequence information and treats the input video as a bag of frames whereas LSTMs use state information to preserve the video sequence.
The DBoF approach with a logistic classification layer produces 2\% (absolute) gains in
Hit@1 and PERR metrics over using simple mean feature pooling and a single-layer
logistic model, which shows the benefits of discrimintatively training a
projection layer to obtain a task-specific video-level representation. The mAP
results for DBoF are slightly worse than mean pooling + logistic model, which we
attribute to slower training and convergence of DBoF on rare classes (mAP is strongly affected
by results on rare classes and the joint class training of DBoF is a disadvantage
for those classes).

The LSTM network generally performs best, except for mAP, where the 1-vs-all
binary MoE classifiers perform better, likely for the same reasons of slower
convergence on rare classes. LSTM does improve on Hit@1 and PERR metrics, as
expected given its ability to learn long-term correlations in the time domain.
Also, in~\cite{beyond}, the authors used data augmentation by sampling multiple
snippets of fixed length from a video and averaged the results, which could
produce even better accuracies than our current results.

We also considered Fisher vectors and VLAD given their recent success in aggregating CNN features at the video-level in
~\cite{xu2015}. However, for the same dimensionality as the video-level representations of the LSTM, DBoF and mean features,
they did not produce competitive results.

\subsubsection{Human Rated Test Set}
We also report results on the human rated test set of over $8000$ videos (see Section~\ref{sec:humangt}) in Table~\ref{tab:humangt}
for the top three approaches. We report PERR, Hit@1, and Hit@5, since the mAP is not reliable given the size of the test set. The
 Hit@1 numbers are uniformly higher for all approaches when compared to the incomplete validation set in Table~\ref{tab:youtube10m}
whereas the PERR numbers are uniformly lower. This is largely attributable to the missing labels in the validation set
(recall of the Validation set labels is around 15\% compared to exhaustive human ratings).
However, the relative ordering of the various approaches is fairly consistent between the two sets, showing that the
validation set results are still reliable enough to compare different approaches.

\subsection{Results on Sports-1M}

Next, we investigate generalization of the video-level features learned using the YouTube-8M dataset and perform transfer learning experiments on the Sports-1M dataset. The Sports-1M dataset~\cite{karpathy2014large} consists of $487$ sports activities with $1.2$ million YouTube videos and is one of the largest benchmarks available for sports/activity recognition. We use the first $360$ seconds of a video sampled at $1$ frame per second for all experiments.

To evaluate transfer learning on this dataset, in one experiment we simply use
the aggregated video-level descriptors, based on the PCA matrix learned on the
YouTube-8M dataset, and train MoE or logistic models on top using target domain training data.

For the LSTM networks, we have two scenarios: 1) we use the PCA transformed features and learn a LSTM model from scratch using these features; or 2) we use the LSTM layers pre-trained on the YouTube-8M task, and fine-tune them on the Sports-1M dataset (along with a new softmax classifier).

Table~\ref{tab:sports1m} shows the evaluation metrics for the various
video-level representations on the Sports-1M dataset.  Our learned features are
competitive on this dataset, with the best approach beating all but the approach
of~\cite{beyond}, which learned directly from the pixels of the videos in the
Sports-1M dataset, including optical flow, and made use of data augmentation
strategies and multiple inferences over several video segments. We also show that
even on such a large dataset ($1$M videos), pre-training on YouTube-8M still helps,
and improves the LSTM performance by $\mathtt{\sim}1$\% on all metrics (vs. no pre-training).

\subsection{Results on ActivityNet}

Our final set of experiments demonstrate the generality of our learned features
for the ActivityNet untrimmed video classification task. Similar to Sports-1M
experiments, we compare directly training on the ActivityNet dataset against
pre-training on YouTube-8M for aggregation based and LSTM approaches. As seen in
Table~\ref{tab:actnet}, all of the transferred features are much better in terms
of all metrics than training on ActivityNet alone. Notably, without the use of
motion information, our best feature is better by up to $80\%$ than the HOG, HOF,
MBH, FC-6, FC-7 features used in~\cite{actnet}. This result shows that features
learned on YouTube-8M generalize very well to other datasets/tasks. We believe
this is because of the diversity and scale of the videos present in YouTube-8M.

\section{Conclusions}
\label{sec:conclusions}
In this paper, we introduce YouTube-8M, a large-scale video benchmark for video classification and representation learning.
With YouTube-8M, our goal is to advance the field of video understanding, similarly to what large-scale image datasets have
done for image understanding. Specifically, we address the two main challenges with large-scale video understanding---(1) collecting
a large \textbf{\em labeled} video dataset, with reasonable quality labels, and (2) \textbf{removing computational barriers} by pre-processing the dataset and providing state-of-the-art frame-level features to build from. We process over 50 years worth of video, and provide features for
nearly $2$ billion frames from more than $8$ million videos, which enables training a reasonable model at this scale within 1 day,
using an open source framework on a single machine! We expect this dataset to level the playing field for academia researchers,
bridge the gap with large-scale labeled video datasets, and significantly accelerate research on video understanding.
We hope this dataset will prove to be a test bed for developing novel video representation learning algorithms,
and especially approaches that deal effectively with noisy or incomplete labels.

As a side effect, we also provide one of the largest and most diverse public visual annotation vocabularies
(consisting of $4800$ visual Knowledge Graph entities), constructed from popularity signals on YouTube
as well as manual curation, and organized into 24 top-level categories.

We provide extensive experiments comparing several strong baselines for video representation learning, including
Deep Networks and LSTMs, on this dataset.
We demonstrate the efficacy of using a fairly unexplored class of models (mixture-of-experts) and show that they can outperform popular classifiers like logistic regression and SVMs. This is particularly true for our large dataset where many classes can be multi-modal. We explore various video-level representations using simple statistics extracted from the frame-level features and model the probability of an entity given the aggregated vector as an MoE. We show that this yields competitive performance compared to more complex approaches (that directly use frame-level information) such as LSTM and DBoF. This also demonstrates that if the underlying frame-level features are strong, the need for more sophisticated video-level modeling techniques is reduced.

Finally, we illustrate the usefulness of the dataset by performing transfer learning
experiments on existing video benchmarks---Sports-1M and ActivityNet. Our experiments show that features learned
on this dataset generalize well on these benchmarks, including setting a new state-of-the-art on ActivityNet.

\bibliographystyle{ieee}
\bibliography{egbib}

\end{document}